\documentclass[10pt,twocolumn,letterpaper]{article}

\usepackage[pagenumbers]{cvpr}      

\usepackage{graphicx}
\usepackage{amsmath}
\usepackage{amssymb}
\usepackage{booktabs}
\usepackage{algorithm2e}
\usepackage{algorithmic}

\makeatletter
\newcommand{\algorithmfootnote}[2][\footnotesize]{%
  \let\old@algocf@finish\@algocf@finish
  \def\@algocf@finish{\old@algocf@finish
    \leavevmode\rlap{\begin{minipage}{\linewidth}
    #1#2
    \end{minipage}}%
  }%
}
\SetKwComment{Comment}{/* }{ */}
 \RestyleAlgo{ruled}

\usepackage[pagebackref,breaklinks,colorlinks]{hyperref}

\usepackage[capitalize]{cleveref}
\crefname{section}{Sec.}{Secs.}
\Crefname{section}{Section}{Sections}
\Crefname{table}{Table}{Tables}
\crefname{table}{Tab.}{Tabs.}

\begin{document}

\title{Anti-Compression Contrastive Facial Forgery Detection }
\author{
Jiajun Huang$^1$,
Xinqi Zhu$^1$, 
Chengbin Du$^{1}$, 
Siqi Ma$^2$,
Surya Nepal$^3$, 
Chang Xu$^1$ 
\\
$^1$School of Computing Science, University of Sydney\\
$^2$School of Engineering and Information Technology, UNSW Canberra, 
$^3$CSIRO Data61\\
\{jhua7177@uni., xinqi.zhu@, chdu5632@uni., c.xu@\}sydney.edu.au,\\
siqi.ma@adfa.edu.au, 
surya.nepal@data61.csiro.au}
\maketitle

\begin{abstract}
Forgery facial images and videos have increased the concern of digital security. It leads to the significant development of detecting forgery data recently. However, the data, especially the videos published on the Internet, are usually compressed with lossy compression algorithms such as H.264. The compressed data could significantly degrade the performance of recent detection algorithms. The existing anti-compression algorithms focus on enhancing the performance in detecting heavily compressed data but less consider the compression adaption to the data from various compression levels. We believe creating a forgery detection model that can handle the data compressed with unknown levels is important. To enhance the performance for such models, we consider the weak compressed and strong compressed data as two views of the original data and they should have similar representation and relationships with other samples. We propose a novel anti-compression forgery detection framework by maintaining closer relations within data under different compression levels. Specifically, the algorithm measures the pair-wise similarity within data as the relations, and forcing the relations of weak and strong compressed data close to each other, thus improving the discriminate power for detecting strong compressed data. To achieve a better strong compressed data relation guided by the less compressed one, we apply video level contrastive learning for weak compressed data, which forces the model to produce similar representations within the same video and far from the negative samples. The experiment results show that the proposed algorithm could boost performance for strong compressed data while improving the accuracy rate when detecting the clean data.
\end{abstract}

\section{Introduction}

\begin{figure}[h]
\centering
  \includegraphics[width=0.48\textwidth]{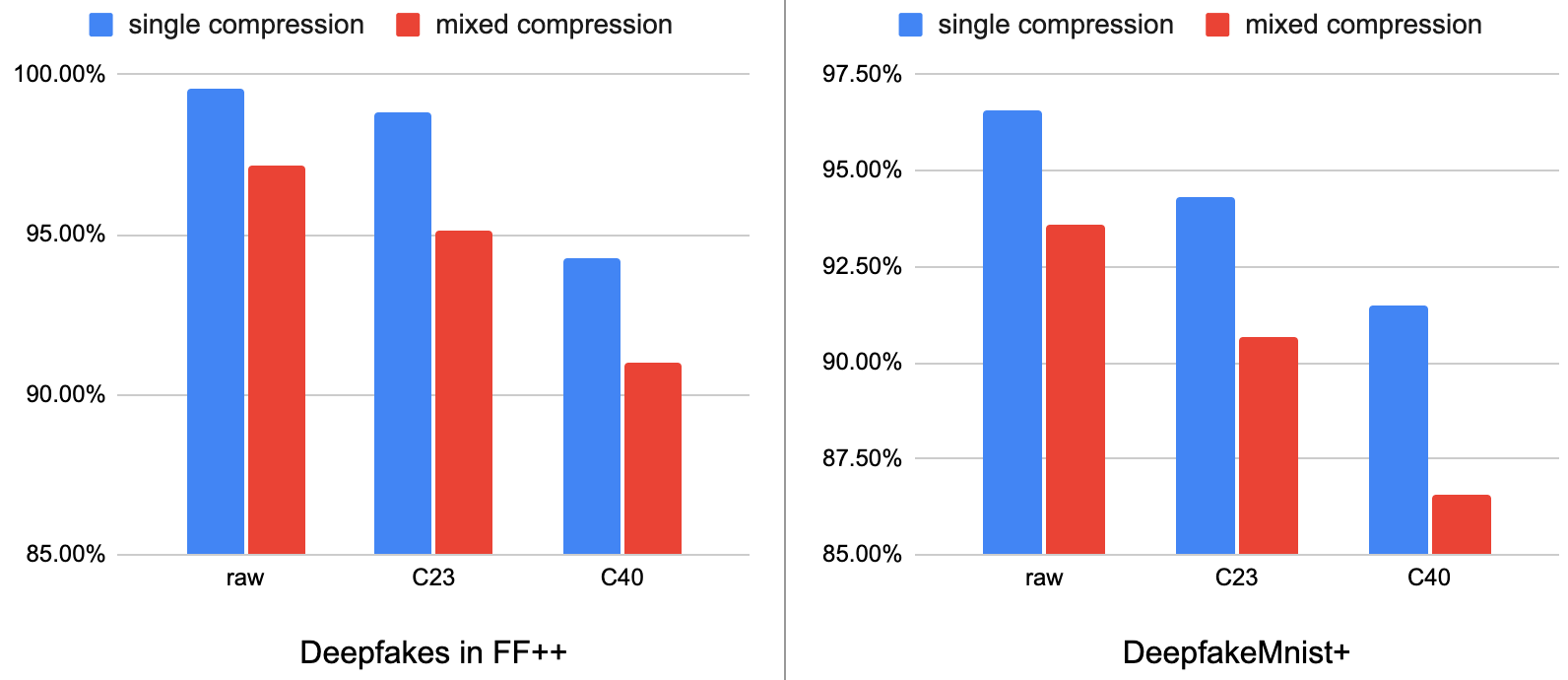}
  \caption{The accuracy performance of models trained with deepfakes in FF++~\cite{roessler2019faceforensicspp} and DeepfakeMnist+~\cite{huang2021deepfake} under different compression level. The \textbf{single compression} means the model trained with data under one specific compression level only. The \textbf{mixed compression} means a single model trained with data in all three different compression level. \textbf{raw} (uncompressed), \textbf{C23} (weak compressed) and \textbf{C40}(strong compressed)}
  \label{fig::video_qualities}
  \vskip -0.3in
\end{figure}
Forgery facial data has become a critical thread for our digital society. The forgery facial data generation techniques can easily change the identity, facial properties, or emotions of a given image or video. The mainstream methods mainly apply Deep Nerual Networks (DNN), especially the Generative Adversarial Network (GAN)~\cite{goodfellow2014generative,chen2020distilling} to achieve this goal, under the name of Deepfake generation. The recent Deepfake methods produce high-quality forgery data that are hard to detect by humans visually, increasing concerns about forgery data since  malicious people could use the techniques to palm off the victims and fake presence and activities.

Many Deepfake methods are proposed to generate visually plausible images or videos by manipulating the identity and facial properties. One major category of Deepfake is face swapping~\cite{GitHubip51:online,GitHubdf58:online,GitHubde97:online,GitHubsh44:online,FakeApp295:online,li2020advancing}, which focuses on replacing the face in one image/video with the face from others. Another category pays more attention to manipulating the facial attributes, including changing expression~\cite{thies2016face2face,thies2019deferred} in images. 
Facial animation methods~\cite{bansal2018recycle,wang2018video,siarohin2020first,suwajanakorn2017synthesizing,burkov2020neural,huang2021deepfake} generate animation videos with a single victim photo and driving video, such that the generated videos present the victim performing the same facial actions as the driving video. These methods can generate forgery data that are hard to detect by human eyes, resulting in severe negative impacts to the victims.

Given the growing anxiety on the high-quality forgery data and its potential negative social impacts, it becomes especially urgent to study the defense techniques against these data. The mainstream detection models rely on the DNN frameworks. The algorithms consider the fogery detection as a binary classification task. By extracting the embeddings of given images or videos, the models predict whether the inputs are real or fake~\cite{roessler2019faceforensicspp, wang2020video}. Some methods explored multiple views of the data \cite{li2017discriminative}, such as facial landmarks~\cite{li2018exposing}, head directions~\cite{yang2019exposing} and facial manipulation masks~\cite{zhao2021multi, li2020face}, while some methods try to detect the inconsistency in the data, such as the image-voice inconsistency\cite{mittal2020emotions} and face-background inconsistency~\cite{zhao2021learning}.

The detection methods have achieved promising performance for discriminating the forgery data. However, video compression could significantly affect the detection performance. The video compression algorithms, such as H.264, are lossy compression algorithms that can estimate some information to reduce the file size. Previous works have shown that the low-quality data can lead to performance degradation caused by domain shift~\cite{roessler2019faceforensicspp, cao2021metric, huang2021deepfake, xu2019positive}. As shown in Figure~\ref{fig::video_qualities}, performance degradation appears when input data are compressed, and this phenomenon deteriorates with even stronger data compression.
Considering that video compression is common when uploading videos to the Internet, developing an algorithm for handling compressed data is important. Several anti-compression methods have been proposed, such as the method with a triplet loss~\cite{kumar2020detecting}, and the GAN-based method that transfers knowledge of manipulation attention from the clean data branch to the compressed branch~\cite{cao2021metric}. However, these methods focus on enhancing the performance on strong compression data but fail to handle the data from other compression levels.

We propose a novel anti-compression forgery facial detection method for handling data from multiple compression levels with contrastive learning. We consider the different compression levels as different views of data, which should share a same internal semantic structure. Instead of matching their representations directly, we expect the model to pay attention to the relations within data in different compression levels. 
Since a model trained on weak-compressed data tends to have better perform than strong-compressed data, we assume its extracted representations reveal more reliable relations for the forgery detection task.
By learning a similar relation between weak and strong compressed data, the model can adapt the relation structure from the weak-compressed data to the strong-compressed data, leading to enhanced performance for detecting strong compressed data.
To ensure the model can produce reliable relations for weak compressed data, we consider the intra frames of one video as the different views. 
The method applies the video-level contrastive learning to achieve a feature extraction that can maintain close distance between intra frames of a video, while enlarging the distance between negative samples.
Our experimental results show that our methods enhance the performance in detecting both clean and compressed data under the mixed compression training manner.

\begin{figure}[]
\centering
  \includegraphics[width=0.47\textwidth]{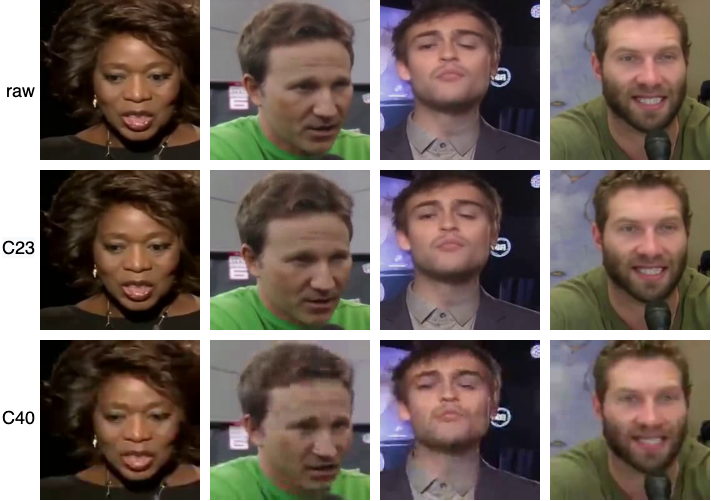}
  \caption{frame samples compressed by H.264 - raw (uncompressed), C23 (weak compressed) and C40 (strong compressed).}
  
 \label{fig::compressed_image}
\vskip -0.2in
\end{figure}

\section{Related Works}
\subsection{Forgery Facial Data and Detection}
The forgery facial data are the images or videos that change the facial identities or attributes of the original data. The recent forgery generation methods, which use DNN and GAN, have achieved high performance such that the generated data are hard to distinguish visually. Identity swapping is the major category that attempts to replace a victim's face in an image or video with another face. The \textit{deepfake}~\cite{GitHubdf58:online} trains two identities-dependent encoder to swap face. Li et al.,~\cite{li2020advancing} proposed an identity-independence framework to generate high-quality results with few samples. RSGAN~\cite{natsume2018rsgan} achieve face-swapping encode the facial information and background information separately. In addition, some methods try to manipulate the facial attributes such as eyes colors, expression, and motion. NeuralTextures~\cite{thies2019deferred} build 3D models of the human face to apply the expression of one person to other. Shao et al.,~\cite{shao2021explicit} proposed an expression transfer method by using adversarial learning to match representation. Huang et al.,~\cite{huang2021deepfake} use an image animation algorithm to generate facial animation data. 

As a respond to high-quality forgery facial data, many forgery detection methods are proposed. \cite{roessler2019faceforensicspp} reveals the DNN could achieve high performance on forgery detection comparing with traditional machine learning algorithms. \cite{wang2020video} uses 3D-CNN framework to detect forgery videos. \cite{guera2018deepfake} implement RNN to extract temporal information within the videos. \cite{rana2020deepfakestack} apply ensemble learning methods to discriminate Deepfake. Instead of using advance DNN framework, some method focus on detecting the inconsistency of the forgery data. \cite{li2020face} detect the unnatural things in the face swapping boundary. \cite{yang2019exposing} highlight the inconsistency between the face and head direction. \cite{mittal2020emotions} consider we can use the gap of facial emotion and voice emotion to detect fake data.

\subsection{Video Compression for Forgery Facial Data}
Video compression is a common method to reduce video size. The compression algorithms, such as H.264, utilize the Discrete Cosine Transform (DCT) to transform video into spectrum domain for quantization operation. The operation will erase the high-frequency information, which is hard to aware by humans, and keep the low-frequency information. In other words, the models lack high-frequency features for discrimination. Information missing could lead to performance degradation. As we can see in Figure~\ref{fig::compressed_image}, it has less difference between raw and weak compressed data(C23). However, the video quality is significantly reduced when we perform strong compression(C40), resulting in blur effects by losing more information. \cite{roessler2019faceforensicspp} shows that the models are hard to achieve high performance with compressed data compared with using clean data as the input. Meanwhile, the compressed data, especially the strong compressed data, could have a significant distribution shift from the clean data~\cite{cao2021metric, huang2021deepfake}. \cite{huang2021deepfake} claims that the models trained with clean data could fail to discriminate strong compressed data. As shown in Table\ref{fig::video_qualities}, the accuracy of detecting uncompressed data is much higher than the compressed data under the single compression training manner. The strong compressed data (C40) could lead to more degradation than the weak compressed one (C23). In addition, the raw data trained model has less performance when detecting the compressed data, which indicates the significant domain shift~\cite{huang2021deepfake}. One naive solution could be training the models with the data under different compression levels together. However, it cannot achieve a similar performance as the single compression training according to Figure~\ref{fig::video_qualities}. The malicious users could compress their forgery videos to remove the forgery signature and achieve a higher spoof rate.

The current anti-compression forgery detection methods focus on implementing metric learning methods to enhance the effect of aligning embeddings. Kumar et al., ~\cite{kumar2020detecting} implement triplet loss~\cite{hoffer2015deep} to enhance the performance for detecting strong compressed Deepfake video. Cao et al., \cite{cao2021metric} apply the two-branch framework to align the manipulated region attention maps within the data under different compression levels. In addition, \cite{qian2020thinking} convert the data into the frequency domain for detection, gaining better discrimination power for strong compressed data. However, the existed methods mainly attempt to improve performance for strong compressed data only. It could be more reasonable to apply a universal model that can handle different compression levels.

\begin{figure*}[h]
  \includegraphics[width=\textwidth]{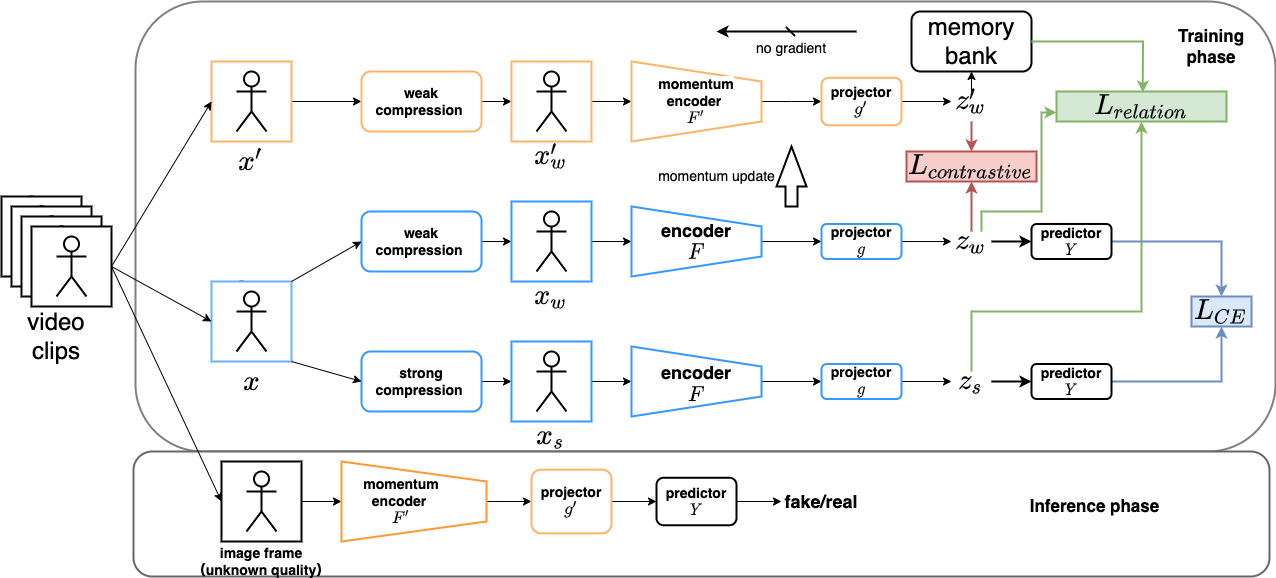}
  \caption{The two-branch framework of our proposed method. In the training phase, the momentum encoder \(F'\) will be updated using \(F\)'s parameter in momentum manner.  We expect that the the relations within embedding of strong compressed data \(z_s\) with a batch of other data should be close to relations of weak compressed data's embedding \(z_w\) with other data. Meanwhile, we enhance weak compressed data relationships with video-level contrastive learning to provide better guideline to learn strong compressed data. In the inference phase, we only keep \(F'\) for prediction.}
  \vskip -0.2in
\end{figure*}
\section{Contrastive Facial Forgery Detection}
Given the videos under two different compression levels (weak and strong compression), we consider that these two kinds of data are two views of the original. The previous methods might directly match their representation or distribution, which could be too strict. We consider the relationships within data could provide more comprehensive information. Due to the high performance in detecting weak compressed data, the relation of weak compressed data might be more precise than strong compressed data. That means approaching weak compressed relationships with the strong compressed relationships could leverage the performance of detecting strong compressed data. Therefore, we proposed a novel method by maintaining similar relations within data under different compression levels to achieve better performance, instead of pushing the representations to be similar directly. To ensure weak compressed relation could provide reliable guideline, the relation should be clear enough. We apply video level contrastive learning to force the representations of weak compressed frames in the same videos close to each other while far away from the data from other classes. 

\subsection{Weak Compression v.s. Strong Compression}
Given a frame \(x\) randomly sampled from a video, we perform weak and strong compression to produce compressed data \(x_w\) and \(x_s\) respectively. A convolutional nerual network (CNN) based encoder \(F\) is implemented to extract features from the input frame images, and a two-layer MLP \(g\) is further applied to produce embedding of extracted features. We calculate the embeddings of \(x_w\) and \(x_s\) through \(z_w = g(F(x_w))\) and \(z_s = g(F(x_s))\), respectively. Consider a batch of anchor samples \(X=\{x_1, x_2, \cdots, x_K\}\), where $x_i$ is the $i$-th anchor randomly selected from the dataset and \(K\) denotes the anchor set size. The embeddings of the anchors can be calculated through encoder $F$ and projector $g$, e.g., \(z_i = g(F(x_i))\) as the embedding of the \(i\)-th anchor \(x\). 

Since the high detection performance on weak compressed data, the model could produce more reliable relations within weak compressed data, e.g., close to the positive samples and far from negative samples. We expect the relations within \(x_s\) and \(X\) should be close to the relations within \(x_w\) and \(X\) to leverage performance for strong compressed data. To achieve this, we first measure the relations within \(x_w\) and \(X\) by calculating the cosine similarity for each anchor embedding \(z_i\):
\begin{equation}
    \text{sim}(z_w, z_i) = \frac{z_w^T z_i}{\left \| z_w \right \|\left \| z_i \right \|}.
\end{equation}
We then can calculate the similarity distribution of \(x_w\) with \(X\) using a SoftMax layer such that:
\begin{equation}
        p^w_i = \frac{\text{exp}(\text{sim}(z_w, z_i)/\tau_w )}{
    \sum^{K}_{k=1}\text{exp}(\text{sim}(z_w, z_k)/\tau_w)},
\end{equation}
where \(\tau_w\) is the SoftMax temperature. Meanwhile, we also measure the similarity distribution of \(x^s\) with \(X\) as well:

\begin{equation}
        p^s_i = \frac{\text{exp}(\text{sim}(z_s, z_i)/\tau_s )}{
    \sum^{K}_{k=1}\text{exp}(\text{sim}(z_s, z_k)/\tau_s)},
\end{equation}
where \(\tau_s\) is another SoftMax temperature different with \(\tau_w\).

For now, we have the two similarity distribution with \(X\) which are consider as the relations under two different compression levels. We achieve similar relations by aligning two similarity distributions with KL divergence as follow:
\begin{equation}
    L_{relation} = D_{KL}(p^w || p^s) = H(p^w, p^s) - H(p^w).
    \label{equ::l_intra}
\end{equation}
As we consider the weak compressed data relations as the guideline, and the strong compressed data relations should approach it. That means, we only update \(p^s\) and fix \(p^w\) with \(L_{relation}\), we can avoid \(H(p^w)\) and only apply for Cross-Entropy loss when calculating \(L_{relation}\). In addition, we can apply sharper \(p^w\), i.e., $\tau_w > \tau_s$), which could highlight the most similar pairs, such that a closer \(p^s\) means it could learn more similar representations between \(x_w\) and \(x_s\). By minimizing \(L_{relation}\), the model could keep closer relationships regardless of the compression level such that enhance the performance for handling inputs with different compression levels. 

\subsection{Real Videos v.s. Fake Videos}
Since we consider weak compressed relation \(p^w\) as the target that guides the model to learn better \(p^s\), it is important to obtain a clear relation for \(p^w\). A clear relation could mean it has a similar representation to its positive pairs and is far from negative pairs. Thus, it is very similar to contrastive learning manner~\cite{chen2020simple, he2019moco}. In this case, we consider the frames within a video as the two views of one video, which is a positive pair. Meanwhile, the frames from the opposite label are treated as negative samples. 

Formally, given frames \(x'\) randomly selected from the same video with \(x\), we perform weak compression to get weak compressed data \(x'_w\). Then we generate the corresponding embedding \(z'_w = g'(F'(x'_w))\) with momentum encoder \(F'\) and momentum projector \(g'\) which have same structures with \(F\) and \(g\) and update the networks using the parameters of \(F\) and \(g\) in momentum manner. After that, we calculate the similarity between \(z'_w\) and \(z_w\), and within \(z'_w\) and the batch of negative samples'embeddings \(U\) (if \(z'_w\) is a real data embedding, then \(U\) contains the forgery data embeddings). Then we minimize the following by applying SoftMax and Cross-Entropy:

\begin{equation}
        L_{contrastive} = -\log(\frac{\text{exp}(\text{sim}(z_w, z'_w)/\tau_v )}{
    \sum_{u\in\{z'_w, U\}}\text{exp}(\text{sim}(z_w, u)/\tau_v)}),
\end{equation}
where \(\tau_v\) is the temperature. By minimizing \(L_{contrastive}\), the representations of intra frames would be close to each other, and they are less similar to the data from the opposite label. It enhances the weak compressed data relation for the guiding model to learn better strong compressed data relation, resulting in better performance when detecting the strong compressed data while improving the detection for weak compressed data.  

\subsection{Optimization}
Behind the representation learning methods above, we also apply two layer linear predictor \(Y\) to allow model learn to classify real and forgery data in a supervised manner with Cross-Entropy loss:
\begin{equation}
    L_{CE} = \text{CE}(Y(z\in\{z_w, z_s\}), \hat{y}),
    \label{equ::l_quality}
\end{equation}
where \(\hat{y}\) is the ground truth label of the given video.
The final loss will be:
\begin{equation}
    L_{final} = L_{CE} + \beta_1*L_{relation} + \beta_2*L_{contrastive},
\end{equation}
where \(\beta_1\) and \(\beta_2\) are the hyper-parameters to adjust the weights of losses.

A larger batch size might be included since we need to measure the relations within multiple samples in \(L_{relation}\) and negative samples in \(L_{contrastive}\), which could be unrealistic for limited GPU capacity. Instead, we maintain the memory bank \(\mathbb{Q}\) of \(K\) pasting samples~\cite{he2019moco}, i.e. \(\{z_k|K=1,2,...,k\}\), and update it with FIFO principle. The bank stores the past embeddings produced by \(F'\), i.e., \(z'_w\), helping us achieve stable performance with smaller batch size. Since we only have two classes in the forgery detection task (either real or forgery), we set up two different banks \(\mathbb{Q}_r\) and \(\mathbb{Q}_f\) for real and forgery samples for easier maintaining. We only use either \(\mathbb{Q}_r\) or \(\mathbb{Q}_f\) for calculating \(L_{contrastive}\) based on the label of current \(x\) and we combine two memory banks for minimizing \(L_{relation}\).

Furthermore, momentum updating is another method to avoid large batch size~\cite{he2019moco, grill2020bootstrap}. Rather than update \(F'\) with backpropagation, we utilize momentum update \(F'\) by using the parameters of \(F\) such that:
\begin{equation}
    F' \leftarrow \beta_m F' + (1-\beta_m)F,
    \label{equ::momentum}
\end{equation}
where \(\beta_m\) is the momentum parameter for updating \(F'\). Similar, we also apply momentum update for momentum projector \(g'\) as well:
\begin{equation}
    g' \leftarrow \beta_m g' + (1-\beta_m)g.
    \label{equ::momentum}
\end{equation}

Unlike previous contrastive learning methods~\cite{he2019moco, grill2020bootstrap, zheng2021ressl}, which keep the encoder F and remove F' after the training, we keep the \(F'\) (and the corresponding \(g'\) and \(Y\)) for the inference phase, which achieves better performance than \(F\).

\begin{algorithm}

\caption{Training on Proposed Method} \label{algorithm:1}
\KwData{ a batch of intra-video frames pairs $x$ and $x'$, target conditions $c$, number of epoch $E$, momentum encoder $F'$ and normal encoder $F$ and their responding projector $g'$ and $g$, predictor $Y$. The real and forgery memory bank: $\mathbb{Q}_r$ and $\mathbb{Q}_s$}.
\KwResult{Well trained $F'$, $g'$ and $Y$ for prediction}
Initial encoder $F$ randomly\;
Copy the initialized parameter of $F$ to $F'$\;
Copy the initialized parameter of $g$ to $g'$\;

\While{network not converge} {
Generate weak compressed data $x_w$ and $x'_w$ and strong compressed data $x_s$\, and calculate corresponding embeddings $z'_w = g'(F'(x'_w))$, $z_w = g(F(x_w))$, $z_s = g(F(x_s))$\;
$\mathbb{Q} = \text{cat}([\mathbb{Q}_r, \mathbb{Q}_f])$,
$p^w=\text{SoftMax}(z_w \mathbb{Q}^T/\tau_t)$, $p^s=\text{SoftMax}(z_s \mathbb{Q}^T/\tau_s)$\;
Calculate $L_{relation}$ with CrossEntropy($p^w$, $p^s$)\;
$logic_{r}$ = cat([$\text{bmm}(z_w^r, z'{_w^r}$), $z^w_r \mathbb{Q}_f^T$]),  $z_r\in real$\;
$logic_{f}$ = cat([$\text{bmm}(z_w^f, z'{_w^f}$), $z^w_f \mathbb{Q}_r^T$]),  $z_f\in fake$\;
Calculate $L_{contrasive} = \text{CrossEntropy}(\text{SoftMax}(logic_{r}/\tau_v), 0) + \text{CrossEntropy}(\text{SoftMax}(logic_{f}/\tau_t), 0)$\;
Calculate $L_{CE}=([Y(z^w_2), Y(z^s_2)], c)$\;
Update $F$, $g$ and $Y$ with $L_{final}$\;
Update $F' = \beta_m F' + (1-\beta_m)F$\;
Update $g' = \beta_m g' + (1-\beta_m)g$\;
Update $Q_r$ and $Q_f$ with $z'_w$
}
\algorithmfootnote{\textbf{cat}: concatenation process, \textbf{bmm}:batch matrix multiplication. When calculating \(L_{contrasive}\), the positive pairs always be the first elements.}

\end{algorithm}

\vspace{-7pt} 
\section{Experiment Results}
In this section, we present empirical studies for our proposed method.
\subsection{Datasets}
Three different forgery facial datasets are used for the experiments which are deepfake dataset from FF++~\cite{roessler2019faceforensicspp}, CeleDF~\cite{Celeb_DF_cvpr20} and DeepfakeMnist+~\cite{huang2021deepfake}. The first two datasets are the popular face-swapping forgery datasets while DeepfakeMinst+ is an action-specific facial animation dataset.

\textbf{Face-swapping datasets}: FF++~\cite{roessler2019faceforensicspp} collects 1,000 Youtube videos that contain human talking scenes. Then they apply \textit{deepfake} algorithm~\cite{GitHubdf58:online} to generate 1,000 facial forgery videos by swapping the faces between two randomly selected videos. Celeb-DF~\cite{Celeb_DF_cvpr20} provides a large forgery and more challenging dataset that contain 590 Youtube videos covering different age and gender groups and generate 5639 face swapping videos.

\textbf{Facial animation dataset}: DeepfakeMnist+~\cite{huang2021deepfake} is a large scale facial animation video dataset. The facial animation tries to manipulate the expression and emotion of the human face, making a face in an image do the same actions as the provided driving video. The dataset contains 10,000 real videos collect from VoxCeleb~\cite{nagrani2017voxceleb}, and generate 10,000 animation videos which ten specific actions such as blinking and nodding (1,000 videos for each action).

As the preprocessing, we apply H.264 lossy compression for each dataset under two different compression ratios - C23 (weak compression) and C40 (strong compression). Then we extract 32 frames for each video and use face recognization model MTCNN~\cite{zhang2016joint} to detect and crop facial regions. Finally, we reshaped the cropped facial images into 224x224 for further training.

\subsection{Baseline Methods}
\textbf{Mixed Compression Training Strategy}: One straightforward way to adapt the compressed data from different levels is to combine these data and train the model. In our experiments, we utilize this strategy to train the encoder \(F\) with cross-entropy loss without any processes for the embedding.

\textbf{L1 loss}: Another straightforward method that might enhance the adaptation could be following the mixed compression training with embedding matching directly. We apply the L1 loss to align the embedding \(z^w\) and \(z^s\) as follow:

\begin{equation}
    L_1 = \sum_i ||z^w_i - z^s_i|| + L_{CE}.
\end{equation}

\textbf{Triplet loss}: \cite{kumar2020detecting} apply triplet loss for single compression training that achieves higher performance with a limited number of data under strong compression. The method extracts an anchor sample \(A\), an embedding of position sample \(P\), and a negative sample \(N\) each time and calculates the triplet loss for these embedding such that:
\begin{align*}
    L_{triplet} &= \max(||g(F(A)) - g(F(P))||^2\\ &- ||g(F(A)) - g(F(N))||^2 + \alpha, 0) + L_{CE},
\end{align*}

where \(\alpha\) is hyper-parameter for the margin. Notice that \(A\), \(P\) and \(N\) are the strong compressed data.

\textbf{Metric learning}: \cite{cao2021metric} apply metric and GAN loss to achieve softer alignment compared with the methods above. The method applies a two-branch network with one trained with weak compressed data while the other trained with strong compressed data. It applies an extra discriminator to align the distribution of the embedding produced by two branches and use the metric loss to separate the outputs between real and forgery inputs. As a result, it enhances the performance of a strong compressed data branch. Please read the original paper for further details.

\textbf{GAN loss}: Inspired by \cite{cao2021metric}, we implement the GAN loss for matching the embedding of two compression levels in mixed compression training manner. Concretely, we apply a 4-linear layers discriminator \(D\) to distinguish the compression level of given embeddings \(z\), and \(F\) should learn to fool \(D\) by producing similar representations under different compression levels. The objective loss will be:
\begin{equation}
L_{GAN} = \log(D(z_w) + \log(1 - D(g(z_s)),
\end{equation}
and the optimization is:
\begin{equation}
\min_{(g, F)}\max_D L_{GAN} + \min_{(F,g,Y)} L_{CE}.
\end{equation}

To be noticed that the previous methods - triplet loss~\cite{kumar2020detecting} and metric learning~\cite{cao2021metric}, utilize the single compression training strategy to train the models. On the other hand, we will perform mixed compression training for L1, GAN loss, and our experiments' proposed method.

\subsection{Training Settings}
We select XceptionNet as the backbone network for encoders \(F'\) and \(F\). It has presented high performance in FF++~\cite{roessler2019faceforensicspp}. We follow the parameters setting of ReSSL~\cite{zheng2021ressl} and MoCo~\cite{he2019moco} for contrastive learning. We set the \(L_{relation}\) temperatures of \(\tau_t = 0.04\) and \(\tau_s = 0.1\), and \(L_{contrastive}\) temperature \(\tau_v = 0.07\). The momentum parameter for updating \(F'\) and \(g'\) is 0.999. In addition, the projectors \(g\) and \(g'\) are two-layer linear projector which output embedding with 512 dimensions and predictor \(Y\) consists of two linear layers as well. Different with previous methods, we apply two separate memory banks \(\mathbb{Q}_r\) and \(\mathbb{Q}_f\) which each has size of (16384, 512). Moreover, we user Adam~\cite{kingma2014adam} optimizer with 0.01 initial learning rate and (0.9, 0.999) for its parameters. For dataset FF++~\cite{roessler2019faceforensicspp} and DeepfakeMnist+~\cite{huang2021deepfake}, we set 0.1 for \(\beta_1\) and \(\beta_2\) for \(L_{final}\), then train the models with 5 epochs in total and the learning rate will be cut into half for every 2 epochs. On the other hand, we perform the warm-up progress for Celeb-DF dataset, by setting \(\beta_1\) and \(\beta_2\) to 0.01 for the first 2,000 steps and adjust them to 1 for the remaining 4,000 steps. We select the best models based on the validation accuracy and we report the performance results in the testing set in the following parts.

\subsection{Detection Performance}

\begin{table*}[t]
\caption{The accuracy performance of models trained in mixed compression manner. From left to right, the first and second columns present the performance on detecting test data under two different compression levels (same with training data). The third column measure the accuracy under raw (uncompressed data). Then we compressed the raw data with JEPG image compression method through Albumentations~\cite{info11020125}, the percentage means the ratio of information remained after compression. We present the accuracy under a range of compression ratios for the last four columns.}
\vspace{-10pt}
\centering
\begin{tabular}{cccccccc}
\toprule
                                 & C23 (weak)                         & C40 (strong)           & 100\% (raw)                        & 75\%                         & 50\%                         & 20\%                         & 10\%    \\
\midrule
\multicolumn{3}{l}{\textbf{Deepfakes - FF++~\cite{roessler2019faceforensicspp}}}                                                             &                              &                              &         \\
\midrule
\multicolumn{1}{c|}{CE loss} & \multicolumn{1}{c|}{95.03\%} & \multicolumn{1}{c|}{91.10\%} & \multicolumn{1}{c|}{94.41\%} & \multicolumn{1}{c|}{94.13\%} & \multicolumn{1}{l|}{92.26\%} & \multicolumn{1}{l|}{88.01\%} & 73.64\% \\
\multicolumn{1}{c|}{CE + L1 loss} & \multicolumn{1}{c|}{94.95\%} & \multicolumn{1}{c|}{91.13\%} & \multicolumn{1}{c|}{94.57\%} & \multicolumn{1}{c|}{94.28\%} & \multicolumn{1}{c|}{92.95\%} & \multicolumn{1}{c|}{88.79} & 74.03\% \\
\multicolumn{1}{c|}{CE + GAN~\cite{cao2021metric}} & \multicolumn{1}{c|}{95.17\%} & \multicolumn{1}{c|}{91.52\%} & \multicolumn{1}{c|}{94.96\%} & \multicolumn{1}{c|}{94.52\%} & \multicolumn{1}{c|}{93.11\%} & \multicolumn{1}{c|}{88.96\%} & 74.15\% \\
\multicolumn{1}{c|}{\textbf{Proposed method}}        & \multicolumn{1}{c|}{\textbf{95.70\%}} & \multicolumn{1}{c|}{\textbf{
93.59\%}} & \multicolumn{1}{c|}{\textbf{95.38\%}} & \multicolumn{1}{c|}{\textbf{\textbf{95.01\%}}} & \multicolumn{1}{l|}{\textbf{94.34\%}} & \multicolumn{1}{l|}{\textbf{91.04\%}} & \textbf{76.19\%} \\
\midrule
\multicolumn{3}{l}{\textbf{DeepfakeMnist+~\cite{huang2021deepfake}}}                                                               &                              &                              &         \\
\midrule
\multicolumn{1}{c|}{CE loss} & \multicolumn{1}{c|}{91.94\%} & \multicolumn{1}{c|}{86.16\%} & \multicolumn{1}{c|}{91.90\%} & \multicolumn{1}{c|}{90.63\%} & \multicolumn{1}{l|}{87.95\%} & \multicolumn{1}{l|}{83.38\%} & 64.78\% \\
\multicolumn{1}{c|}{CE + L1 loss} & \multicolumn{1}{c|}{92.23\%} & \multicolumn{1}{c|}{86.32\%} & \multicolumn{1}{c|}{92.39\%} & \multicolumn{1}{c|}{90.71\%} & \multicolumn{1}{c|}{87.35\%} & \multicolumn{1}{c|}{83.32} & 64.83\% \\
\multicolumn{1}{c|}{CE + GAN~\cite{cao2021metric}} & \multicolumn{1}{c|}{92.16\%} & \multicolumn{1}{c|}{86.53\%} & \multicolumn{1}{c|}{93.04\%} & \multicolumn{1}{c|}{90.52\%} & \multicolumn{1}{c|}{88.71\%} & \multicolumn{1}{c|}{83.74\%} & 65.28\% \\
\multicolumn{1}{c|}{\textbf{Proposed method}}        & \multicolumn{1}{c|}{\textbf{93.10\%}} & \multicolumn{1}{c|}{\textbf{87.48\%}} & \multicolumn{1}{c|}{\textbf{93.38\%}} & \multicolumn{1}{c|}{\textbf{92.63\%}} & \multicolumn{1}{l|}{\textbf{89.13\%}} & \multicolumn{1}{l|}{\textbf{84.32\%}} & \textbf{66.00\%} \\
\midrule
\multicolumn{3}{l}{\textbf{CelebDF~\cite{Celeb_DF_cvpr20}}}                                                                      &                              &                              &         \\
\midrule
\multicolumn{1}{c|}{CE loss} & \multicolumn{1}{c|}{88.87\%} & \multicolumn{1}{c|}{82.92\%} & \multicolumn{1}{c|}{88.49\%} & \multicolumn{1}{c|}{88.61\%} & \multicolumn{1}{l|}{87.67\%} & \multicolumn{1}{l|}{84.14\%} & 65.39\% \\
\multicolumn{1}{c|}{CE + L1 loss} & \multicolumn{1}{c|}{89.28\%} & \multicolumn{1}{c|}{83.15\%} & \multicolumn{1}{c|}{88.35\%} & \multicolumn{1}{c|}{88.94\%} & \multicolumn{1}{c|}{87.91\%} & \multicolumn{1}{c|}{84.23} & 65.58\% \\
\multicolumn{1}{c|}{CE + GAN~\cite{cao2021metric}} & \multicolumn{1}{c|}{89.54\%} & \multicolumn{1}{c|}{83.22\%} & \multicolumn{1}{c|}{89.27\%} & \multicolumn{1}{c|}{88.70\%} & \multicolumn{1}{c|}{88.14\%} & \multicolumn{1}{c|}{84.76\%} & 66.37\% \\
\multicolumn{1}{c|}{\textbf{Proposed method}}        & \multicolumn{1}{c|}{\textbf{90.70\%}} & \multicolumn{1}{c|}{\textbf{83.74\%}} & \multicolumn{1}{c|}{\textbf{90.23\%}} & \multicolumn{1}{c|}{\textbf{89.68\%}} & \multicolumn{1}{l|}{\textbf{88.05\%}}  & \multicolumn{1}{l|}{\textbf{85.45\%}} & \textbf{68.71\%} \\
\bottomrule
\end{tabular}
\label{table::mixed_quality}\vskip -0.2in
\end{table*}
We first compare the models trained with mixed compression manner. Table~\ref{table::mixed_quality} shows the performance of training the models in the mixed compression manner. The embedding matching methods (L1 and GAN~\cite{cao2021metric}) could enhance the detection accuracy in both two compression level. In addition, the GAN loss, which provides a softer matching constraint, might provide better improvement. Our proposed method shows further improvements in all three different datasets compared with the other two matching methods, which improve 2.07\%, 0.95\%, and 0.52\% accuracy on strong compressed data than GAN loss. Furthermore, we also compress the data into multiple levels to evaluate the adaptation of unknown compression level. We apply the JEPG image compression algorithm through Albumentations~\cite{info11020125} to generate compressed data with five different levels based on the raw video data. The percentages indicate the ratio of the remaining information. According to Table~\ref{table::mixed_quality}, the models trained with L1 and GAN loss present similar performance in detecting data in compression levels. At the same time, our proposed method achieves the best performance around these methods in all three datasets. In addition, the method presented significant improvement under the extremely compressed data (10\% information remained), which increased 2.55\%, 1.22\%, and 3.32\% compared with the one only using CE loss.

Meanwhile, we compare the models trained with the single compression manner, that is, training the models with either C23 or C40 compression ratio. As Table~\ref{table::single_performance} shows, only using cross-entropy loss for training in one single might be failed to adapt the data from another compression level. It is interesting to notice that the strong compressed data could provide more generalization in face-swapping datasets. One reason could be that the discriminate information of strong compression is a subset of the weak compressed one. The discriminate power learned from strong compressed data could also be used for detecting the weak compressed one. On the other hand, the previous methods which present improvement on detecting strong compressed data cannot achieve high performance on weak compressed data. The metric learning~\cite{cao2021metric} provides high accuracy in weak compressed data than the triplet loss~\cite{kumar2020detecting} since they utilize a weak compression branch to to guide the learning of the strong compression branch. Finally, the proposed method, which is trained with mixed compression training, provides better adaptation to different compression levels. Although it has slight degradation compared with the models trained with strong compressed data, it has around 7\% improvement on detecting weak compressed data.

\begin{table}[]
\centering
\caption{The accuracy of models trained with single compression manner and compare with our proposed method. We use XceptionNet for all experiments. }
\vspace{-10pt}
\begin{tabular}{cccc}
\toprule
                                                    & C23                          & C40   & AVG  \\
                                                    \midrule
\multicolumn{3}{l}{\textbf{Deepfakes in FF++~\cite{roessler2019faceforensicspp}}}                                                        \\
\midrule
\multicolumn{1}{c|}{CE loss (C23)}               & \multicolumn{1}{c|}{98.89\%} & \multicolumn{1}{c|}{73.86} & 86.38\% \\
\multicolumn{1}{c|}{CE loss (C40)}               & \multicolumn{1}{c|}{86.85\%} & \multicolumn{1}{c|}{94.21\%} & 90.53\% \\
\multicolumn{1}{c|}{CE + triplet loss (C23)~\cite{kumar2020detecting}}     & \multicolumn{1}{c|}{99.42\%} & \multicolumn{1}{c|}{76.89\%} & 88.16\%\\
\multicolumn{1}{c|}{CE + triplet loss (C40)~\cite{kumar2020detecting}}     & \multicolumn{1}{c|}{85.71\%} & \multicolumn{1}{c|}{94.46\%} & 90.09\%\\
\multicolumn{1}{c|}{metric learning (C23)~\cite{cao2021metric}} & \multicolumn{1}{c|}{97.58\%} & \multicolumn{1}{c|}{80.53\%} & 89.06\%\\
\multicolumn{1}{c|}{metric learning (C40)~\cite{cao2021metric}} & \multicolumn{1}{c|}{88.62\%} & \multicolumn{1}{c|}{94.71\%} & 91.67\%\\
\multicolumn{1}{c|}{\textbf{Proposed method}}                & \multicolumn{1}{c|}{95.70\%} & \multicolumn{1}{c|}{93.59\%} & \textbf{94.65}\%\\
\midrule
\multicolumn{3}{l}{\textbf{DeepfakeMnist+~\cite{huang2021deepfake}}}                                                           \\
\midrule
\multicolumn{1}{c|}{CE loss (C23)}               & \multicolumn{1}{c|}{94.37\%} & \multicolumn{1}{c|}{82.98\%} & 88.68\%\\
\multicolumn{1}{c|}{CE loss (C40)}               & \multicolumn{1}{c|}{76.69\%} & \multicolumn{1}{c|}{91.27\%} & 83.98\%\\
\multicolumn{1}{c|}{CE + triplet loss (C23)~\cite{kumar2020detecting}}     & \multicolumn{1}{c|}{94.63\%} & \multicolumn{1}{c|}{81.28\%} &  87.96\%\\
\multicolumn{1}{c|}{CE + triplet loss (C40)~\cite{kumar2020detecting}}     & \multicolumn{1}{c|}{77.16\%} & \multicolumn{1}{c|}{91.53\%} & 84.35\%\\
\multicolumn{1}{c|}{metric learning (C23)~\cite{cao2021metric}} & \multicolumn{1}{c|}{94.41\%} & \multicolumn{1}{c|}{84.70\%} & 89.56\%\\
\multicolumn{1}{c|}{metric learning (C40)~\cite{cao2021metric}} & \multicolumn{1}{c|}{76.69\%} & \multicolumn{1}{c|}{91.67\%} & 84.18\%\\
\multicolumn{1}{c|}{\textbf{Proposed method}}                & \multicolumn{1}{c|}{93.38\%} & \multicolumn{1}{c|}{87.48\%} & \textbf{90.43}\%\\
\midrule
\multicolumn{3}{l}{\textbf{CelebDF~\cite{Celeb_DF_cvpr20}}}                                                                  \\
\midrule
\multicolumn{1}{c|}{CE loss (C23)}               & \multicolumn{1}{c|}{93.07\%} & \multicolumn{1}{c|}{55.09\%} & 74.08\%\\
\multicolumn{1}{c|}{CE loss (C40)}               & \multicolumn{1}{c|}{79.43\%} & \multicolumn{1}{c|}{84.49\%} &  81.96\%\\
\multicolumn{1}{c|}{CE + triplet loss (C23)~\cite{kumar2020detecting}}     & \multicolumn{1}{c|}{94.25\%} & \multicolumn{1}{c|}{57.83\%} & 76.04\%\\
\multicolumn{1}{c|}{CE + triplet loss (C40)~\cite{kumar2020detecting}}     & \multicolumn{1}{c|}{80.52\%} & \multicolumn{1}{c|}{84.83\%} & 82.67\%\\
\multicolumn{1}{c|}{metric learning (C23)~\cite{cao2021metric}} & \multicolumn{1}{c|}{92.87\%} & \multicolumn{1}{c|}{65.38\%} &  79.13\%\\
\multicolumn{1}{c|}{metric learning (C40)~\cite{cao2021metric}} & \multicolumn{1}{c|}{82.02\%} & \multicolumn{1}{c|}{84.90\%} &  83.46\%\\
\multicolumn{1}{c|}{\textbf{Proposed method}}                & \multicolumn{1}{c|}{90.70\%} & \multicolumn{1}{c|}{83.79\%}   & \textbf{87.26}\%\\
\bottomrule
\end{tabular}
\label{table::single_performance}
\vskip -0.15in
\end{table}

\subsection{Ablation study for proposed method}
\begin{table}[]
\caption{The ablation study of proposed method tested with deepfake in FF++~\cite{roessler2019faceforensicspp}. The performances are measured by accuracy.}
\vskip -0.1in
\centering
\begin{tabular}{ccc}
\toprule
                                                                                               & C23                          & C40     \\
\midrule

\multicolumn{1}{c|}{\begin{tabular}[c]{@{}c@{}}CE loss\end{tabular}} & \multicolumn{1}{c|}{95.03\%} & 91.10\% \\
\multicolumn{1}{c|}{CE loss + momentum update only} & \multicolumn{1}{c|}{95.21\%} & 92.26\% \\
\multicolumn{1}{c|}{CE loss + \(L_{contrastive}\)}                                                        & \multicolumn{1}{c|}{95.67\%} & 92.63\% \\
\multicolumn{1}{c|}{CE loss + \(L_{relation}\)}                                                      & \multicolumn{1}{c|}{95.24\%} & 93.07\%\\
\multicolumn{1}{c|}{CE loss + \(L_{relation}\) + \(L_{contrastive}\)}                                                      & \multicolumn{1}{c|}{95.70\%} & 93.59\%\\

\multicolumn{1}{c|}{\begin{tabular}[c]{@{}c@{}}CE loss + \(L_{relation}\) + \(L_{contrastive}\)\\ trained with raw and C40\end{tabular}}                                                      & \multicolumn{1}{c|}{94.02\%} & 91.49\%\\
\bottomrule
\end{tabular}
\label{table::ablation}
\vskip -0.25in
\end{table}

In this part, we perform ablation study to explore the benefit of using each module of our proposed method, including momentum update, \(L_{relation}\) and \(L_{contrastive}\). We measure the performance using the deepfake dataset in FF++~\cite{roessler2019faceforensicspp}.  

\textbf{The benefit of momentum update}: According to Table~\ref{table::ablation}, we can notice the momentum update for \(F'\) have provided significant improvement comparing with the normal gradient update in \(F\). It has a 0.18\% in weak compressed data, and more increasing in strong compressed data with a 1.15\% gap. The results could indicate that the training of forgery detection are suffered the over-fitting problem in the mixed compression training manner. And momentum update could provide more stable update and approach to higher performance.

\textbf{The benefit of video contrastive matching}: As Table~\ref{table::ablation} shows, the video contrastive loss enhances the performance on detecting weak compressed data, with a further 0.46\% accuracy improvement than the momentum update and 0.64\% increment compared with using CE loss. In addition, it also provides enhances the detection for strong compressed data with a 0.37\% improvement than the momentum update. The increased accuracy means the model could learn better representations with closer relations within intra-frames and a larger gap with the negative samples. And the improvement also helps the network on detecting strong compressed data.

\textbf{The benefit of compression relations matching}: The compression relation loss could engage a higher accuracy rate on detecting strong compressed data as presented in Table~\ref{table::ablation}, with 0.81\% and 1.97\% compared with momentum update and normal update. However, as we only consider the weak compressed data as the target and fix them when updating the loss, it has a minor improvement for the weak compressed data.

\textbf{The benefit of combining two losses}: With combining all proposed loss and momentum update, the performance is further improved in detecting both weak and strong compressed data, which achieve 95.70\% and 93.59\%, respectively. It increases 0.52\% of accuracy compared with apply \(L_{relation}\) only, indicating the more clear weak compressed relation could provide a better guideline for aligning strong compression relation.

\textbf{The impact of training compression level} In our proposed method, we consider the weak compressed data as the guideline to improve the detection accuracy on strong compressed data. We also explore the situation that we use raw data rather than weak compressed data. As shown in Table~\ref{table::ablation}, we notice that it provides worse performance, which has 1.01\% degradation compared with the normal training with CE loss. It could indicate that exists a small domain shift between raw and weak compressed data. In addition, it also means it has a larger distribution gap with the strong compressed data, resulting in a minor improvement in detecting strong compressed data.

\begin{table}[]
\caption{The accuracy performance of models trained with different size of memory bank \(\mathbb{Q}_r\) and \(\mathbb{Q}_f\) in deepfake of FF++~\cite{roessler2019faceforensicspp}.}
\vskip -0.15in
\begin{tabular}{cccccc}
\toprule
                         & 256                          & 1024                         & 4096                         & 16384                                 & 32768   \\
\midrule
\multicolumn{1}{c|}{C23} & \multicolumn{1}{c|}{95.31\%} & \multicolumn{1}{c|}{95.49\%} & \multicolumn{1}{c|}{95.53\%} & \multicolumn{1}{c|}{\textbf{95.70\%}} & 95.63\% \\
\multicolumn{1}{c|}{C40} & \multicolumn{1}{c|}{92.82\%} & \multicolumn{1}{c|}{93.05\%} & \multicolumn{1}{c|}{93.27\%} & \multicolumn{1}{c|}{\textbf{93.56\%}} & 93.46\%\\
\bottomrule
\end{tabular}
\label{table::bank_size}
\vskip -0.3in
\end{table}
\textbf{The impact of memory size}: We also explore how the sizes of memory banks \(\mathbb{Q}_r\) and \(\mathbb{Q}_f\) could affect the performance and Table~\ref{table::bank_size} present the results. The results indicate that the \(L_{relation}\) is more sensitive with the size than \(L_{video}\). When the memory sizes are relatively small, it could cause more negative impact to the performance of C40 than C23 data. One explanation could be that learning a similar relation requires a larger batch of embeddings, while it has less requirement for matching embeddings and pushing away from the negative samples. In addition, a larger batch size could improve the detection performance, but the further increase of batch size could also affect the performance. 
\vspace{-5pt} 
\section{Conclusion}
\vspace{-5pt} 

In conclusion, we proposed a novel forgery facial detection method to handle data in different compression levels with contrastive learning. The method enhances the performance in detecting both weak and strong compressed data by learning better relations within weak compressed data and maintaining similar relations under different compression levels. Furthermore, our experiment results and ablation study show that the proposed method adapt multiple compression levels better than previous methods.

{\small
\bibliographystyle{ieee_fullname}
\bibliography{egbib}
}

\end{document}